\def\year{2019}\relax
\documentclass[letterpaper]{article} 
\usepackage{aaai19}  
\usepackage{times}  
\usepackage{helvet} 
\usepackage{courier}  
\usepackage[hyphens]{url}  
\usepackage{graphicx} 
\def\UrlFont{\rm}  
\usepackage{graphicx}  
\usepackage{lipsum}
\usepackage{multicol}
\usepackage{xcolor} 

\title{Evaluation of a Recommender System for Assisting Novice Game Designers}
\author{\Large \textbf{} \\ \Large \textbf{Tiago Machado, Daniel Gopstein, Angela Wang, Oded Nov, \textsuperscript{\rm *}Andrew Nealen, Julian Togelius}\\ 
New York University\\
\textsuperscript{\rm *} University of Southern California\\
\{tiago.machado, dgopstein, awayc, onov, julian.togelius\} at nyu.edu \\
\textsuperscript{\rm*}\{anealen\} at cinema.usc.edu
}
 
\begin{document}

\maketitle

\begin{abstract}
Game development is a complex task involving multiple
disciplines  and  technologies.
Developers and researchers alike have suggested that AI-driven game design assistants may improve developer workflow. 
We
present a recommender system for assisting humans in game
design as well as a rigorous human subjects study to validate it.  The AI-driven game design assistance system suggests game mechanics to designers based on characteristics of the game
being developed.  We believe this method can
bring creative insights and increase user’s productivity.  We
conducted quantitative studies that showed the recommender system increases
users' levels of accuracy and computational affect, and decreases their levels
of workload. 
\end{abstract}

\section{Introduction}
Game development is a constantly evolving field. With time, game productions tends to be larger and more complex~\cite{Blow:2004:GDH:971564.971590}. 
This rising complexity calls for more capable tools, and consequently there is constant development of game development toolchains. 
However, there are those who argue that game development tools could be so much more, given available technology~\cite{Kasurinen:2013:GDE:2460999.2461004}. 
In particular, various types of artificial intelligence could be used to assist designers. There exist various prototypes of AI-driven game design assistance tools in the literature, but these tools are generally tied to a particular game and are rarely subject to rigorous user studies to evaluate their human factors.
In this paper we present a formal evaluation of Pitako, a game design recommendation system~\cite{machado2019pitako}, which is part of an AI-driven game design assistance tool named Cicero~\cite{Machado2018AIAssistedGD}. Pitako assists novice designers through a recommender system which explores a whole library of games, and suggests game mechanics that have a variety level of confidence matching with the ones being designed. By providing such suggestions, Pitako performs an automatic exploration of possible designs and provides users an easy way to play with pieces of different games and explore their creativity\cite{Shneiderman:2007:CST:1323688.1323689}. 
We briefly present Pitako, and then report the results of a quantitative study to show how users react in presence of an algorithm experience (AE) as described by \cite{Oh:2017:UVT:3025453.3025539}. Eighty-seven (87) participants were divided into two groups, both had to design the same game, however one of the groups was instructed to use the AI-assistant. In particular we evaluated how users' level of self-efficacy, affect, workload and accuracy varied for each group. Our hypotheses were that the presence of an AI-driven game design assistance tool would \textbf{(H1)} decrease the workload, \textbf{(H2)} increase self-efficacy, \textbf{(H3)} increase affect, and \textbf{(H4)} increase accuracy. By having appropriate tools the users can think more about their design and spend less time dealing with tool issues\cite{Shneiderman:2007:CST:1323688.1323689}. We hope the results presented here will foster a new generation of AI-assistants for game design tasks focused both on its technical aspects and the human ones. This formal approach about evaluating an AI-game design tool is the main contribution of this paper. 


\section{Related Work}
This section details work that inspired this paper, for a better understanding we divided it into three categories: AI-Game Design Assistants, User studies for AI-game design assistants, and Game Tools Issues.




\subsection{AI-Game Design Assistants}
A human-in-the-loop process is essential for AI-game design assistants, here we list some examples that we consider worth to note. Tanagra \cite{smith2010tanagra} is a level generator for 2D platform games.  It works as a mixed-initiative tool where a human and an algorithm works collaborate  to design level. The level generator responds in realtime to any input given by the user. It supports changes in the geometry level and at the level's pace through a timeline. Ropossom \cite{shaker2013ropossum} is also a level generator. However its targets are puzzles game, particularly in this case, Cut the Rope. The system brings two modules, one based on evolutionary computing techniques to generate the levels and another one with an agent, based on search algorithms, that guarantees the generated level is playable. Also using an evolutionary approach, \cite{6185648} designed a spaceship generator. By providing a list (population) of spaceships, the system iteratively changes the appearance of the new populations based on the user selection. Evolutionary algorithms are also the foundation of Sentient Sketchbook \cite{liapis2013sentient}. The tool is a suggestion engine for map design. It updates its recommendations based on the user interactions with the system in realtime. The whole concept works like a CAD (Computer Aided Design) tool. Having the user's map as input, the system automatically suggests navigation paths and guarantees playability.    

\subsection{User Studies For AI-driven Game Design Assistants}
There is no silver bullet in game or software development. At the same time as a new technology solves old complexities, new complexities may appear \cite{4420077}. AI-driven game design assistants are no exception to this rule, making it important that such tools are rigorously evaluated. Previously, the debugging functionality of the Cicero system was evaluated using a quantitative approach \cite{Machado2018AIAssistedGD}. That study compared how humans perform when debugging a game with AI assistance and without it, and found that the AI assistance system offered significant benefit. Another investigation of the benefits of AI-driven game assistants can be found in the work by  \cite{alvarez2018fostering}, in which the authors analyze dungeon design levels and discuss the cost reduction and creativity gains such tools can bring. Finally, Morai Maker is a tool designed to evaluate how machine learning might influence designers when creating levels for Super Mario Bros \cite{guzdial2019friend}. 

\subsection{Game Tool Issues}
Researchers and professionals state that the tools used to implement the games is among one of the problems that lead  projects to delays, cancellations and failures \cite{petrillo2009went,washburn2016went}. 
In a qualitative study,  \cite{Kasurinen:2013:GDE:2460999.2461004} interviewed 27 developers from seven game startup companies and one of the conclusions is that developers feel the lack of tools more adaptable to their needs. They cited to go off game engines often when they need to prototype and test ideas. Another issues are the learning curve of these tools, it may slope and slow down the progress \cite{Kasurinen:2013:GDE:2460999.2461004}, there is also little code reuse between and within games, and lack of studies about the efficacy of these tools \cite{murphy2014cowboys}.
Motivated by the issues presented in this section, we used Pitako's automatic exploratory design features to evaluate how designers' self-perception of their tasks change in presence of an algorithm. To address the lack of formal user studies, we evaluated Pitako across three dimensions: workload, affect, self-efficacy, and accuracy.

\section{System Design}
The foundations of the Pitako recommender system are outlined in this section; it is described in more depth in~\cite{machado2019pitako}.
\subsection{GVGAI \& VGDL}
The system was implemented by Machado et al as part of the Cicero Game Design Assistance System~\cite{Machado2018AIAssistedGD}, which itself is built on top of the General Video Game Framework (GVGAI) \cite{perez20162014} and the Video Game Description Language (VGDL) \cite{schaul2013video,ebner2013towards}. 
This framework was used because it allows the design of methods that can be applied across different game genres. Most frameworks for AI-game tools like the Mario AI framework \footnote{http://julian.togelius.com/Togelius2010The.pdf}, Pommerman\footnote{https://arxiv.org/abs/1809.07124}, and Angry Birds AI\footnote{https://aibirds.org/} only work for one game. GVGAI has more than a hundred games available, in genres ranging from action, to puzzles, to space shooters. They are always in 2D and look like Atari 2600 games. While the system has shortcomings in representational power and usability, it provides a good amount of generality within a domain.
The GVGAI framework has an associated competition that runs annually \cite{perez20162014}. Developers and researchers can participate by submitting their algorithms to one of the many categories available like game play agents, level generators, game generators, and two-player game agents.
The games of the framework are written in VGDL. The language is simple and human-readable. It defines all its games in four sets, the most important ones for this study are:
the game element set, which describes the behaviors of each game element (including the player, the enemy, an obstacle, a power-up, etc), 
and the game interaction set, which defines what happens when two elements collide (interact) with each other.

\subsection{Frequent itemset data mining}
The games available in the GVGAI framework were all designed by humans. The game element set and the interaction set presents elements which are associated to each other. By distilling the two sets, they look like transaction lists like the ones present in frequent item set data mining algorithms, for example, the Apriori algorithm \cite{guo2017application}. The idea of the algorithm is to find relevant association rules in large databases containing lots of transactions. An association rule shows how the presence of an item \textbf{A} implies in the presence of an item \textbf{B}. This implication comes with a level of confidence, i.e., the probability that \textbf{B} will appear if \textbf{A} is already contained in the transaction. A common example comes from Market Basket Analysis \cite{brin1997dynamic} and a statement that 90\% of transactions which contains bread and butter, also will contain milk \cite{agrawal1993mining}. 

\subsection{Catalog}
The basis of the recommender system is a catalog, where every game element set and every interaction set is stored as a list of transactions. Therefore each game element of an specific type is coded as an individual number, the same is done for the transactions. The final result is two tables: one whose the entries are games and its elements coded into numbers; another whose entries are tuples of game elements type and the interaction they fire when colliding. These tables work like a catalog of game elements. By breaking down every game description to the level of its individual parts it facilitates the task of generating the input for the Apriori algorithm. It also makes easier the exchange of elements and information from one game to another.

\begin{figure*}[ht]
  \includegraphics[width=\linewidth]{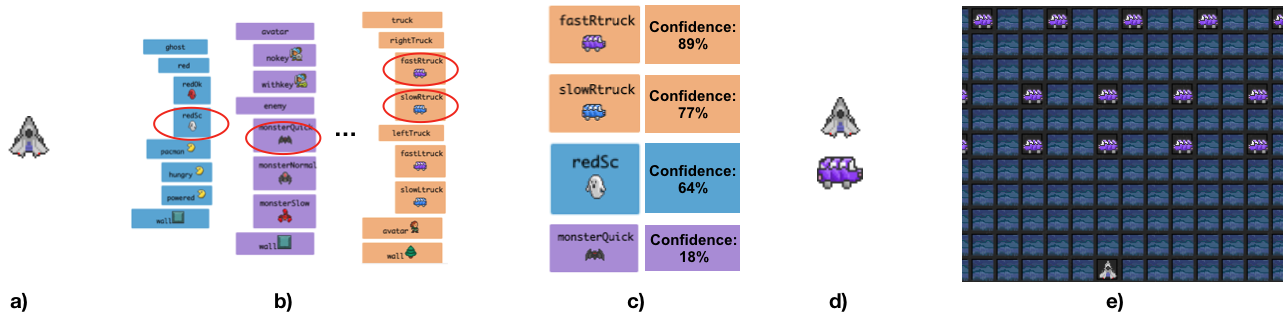}
  \caption{The recommendation process starts by getting the designer game description set. In \textbf{a)} the user has only one game element. In \textbf{b)}, the system is performing a frequent itemset data mining algorithm to identify association(s) that matches the element in the designer's game to elements in the games' catalog. The red ellipses represent elements from different games which are candidates for being recommended. This procedure yields a list of recommendations sorted in descending order by the rule's confidence level \textbf{c}. The designer picks the recommendation with the highest confidence level \textbf{d}. Then, starts the to use it as a new element to the game. This element came from a clone of the famous game \textit{Frogger}. And in this example, the user is mixing elements from \textit{Frogger} and \textit{Space Invaders}.}
  \label{process}
\end{figure*}

\subsection{Generating Recommendations}
With the raw data of the game descriptions already converted to the kind of data necessary to run the Apriori algorithm, everything is set up for generating recommendations. The output of the algorithm is a list of association rules for game elements and another one for game interactions. All this process is done offline so the association rules can be available when the user is designing a game. Therefore, whenever a user adds or remove a game element, the user element set will be compared to the list of association rule. The result of this comparison is a list of elements which sign positive for a match with the user's set.

It has been argued that recommender systems can be inimical to creativity, serving as ``Weapons of Math Destruction'' as they take away from genuine user choice~\cite{o2017weapons}. However, the Pitako system is designed to avoid this by offering multiple suggestions.  The list presented to the user are associated with various levels of confidence. Therefore, it is up to the user to choose in which direction the algorithm will work. In case the goal is to design a clone of an existing game, selecting the highest level recommendations is the appropriate choice. In case the goal is to find new alternatives, the lower confidence recommendations are the way to go. To avoid redundancy we will not describe the interaction recommendation process, which is similar to the one already described.



\begin{figure*}[ht]
  \includegraphics[width=\linewidth]{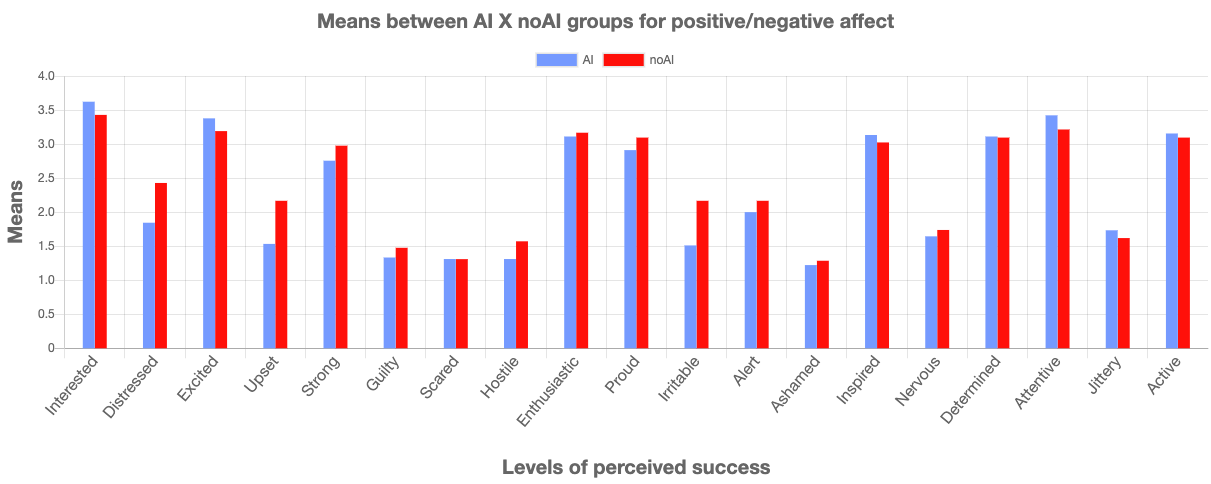}
  \caption{Means for the Positve/Negative affect from the two groups observed (AI and non-AI) in this experiment.}
  \label{panas}
\end{figure*}

\section{User Study}
A quantitative research design was used including three standardized forms and a procedure. 

\subsection{Subjects}
Eighty-seven (87) students from our university computer science and interactive media departments participated in the study. They were recruited by their departments email newsletter and had to fill a preliminary form to be contacted later in case of agreement in participating of the experiment. The recruitment message informed them that experience with authoring tools was required as well as an interest in playing games. The participants were rewarded for volunteering in this study with a 20 USD Amazon Gift Card.

\subsection{Research Material}
Two versions of Pitako (one with the recommender system and another one without it) were made available to download by the participants. For each software, a video tutorial was available as well. 

\subsection{Sources of Data}
Four sources of data were used to evaluate the effectiveness of the two versions of the system. The NASA-TLX \cite{hart1988development} was used to evaluate the participants' perceived workload in all the dimensions of the questionnaire that covers from mental to physical effort. The PANAS \cite{crawford2004positive} was used to measure participants' computtional affect. It is a 20-item measure of positive affect (10 items) and negative affect (10 items). All items are rated on a Likert-scale ranging from 1 (not at all) to 5 (extremely). Subjects responded to PANAS items based on to what extent they feel after completing the task. Computer self-efficacy was used to measure how confident users were about their skills with respect to the evaluated systems \cite{marakas1998multilevel}. The last source of data were the games designed by the users during the experiment. We used them to evaluate how accurate were users in design \textit{Space Invaders} with the recommender system and without it.

\subsection{Procedure}
Once the participants agreed to be part of the experiment, an email was sent to them. The email contained a link in which they could download the software and a document to guide them about how to do the study. In the first part of the experiment, the guide explained the users how to install the software. Following, the guide explained the task to be executed: design \textit{Space Invaders} without using the recommender system or design \textit{Space Invaders} by using the recommender system. A video showing a prototype similar to the one to be designed was linked in the guide document. Therefore, those not familiar with \textit{Space Invaders} could have a feel about how the mechanics to be implemented works. 
At this point, the guide provided video tutorial links and instructions to the participants about each one of the components they had to learn and follow to perform the task. In both versions, participants were timed in 30 minutes after pressing the confirming button to start the task.
After the experiment was over, either by decision of the users, or by time, they were prompted to fill the forms (NASA-TLX, PANAS, and Computer Self-Efficacy) and submit their game.

\section{Results}

In this section we will present our results first by listing in the following subsections the hypothesis that our statistical analysis showed significance: workload, affect, and accuracy; and then the results for computational self-efficacy. Please note that we divided the participants in two groups. The AI group have the participants who have done the experiment with Pitako (recommender system), while the the noAI group have done the experiment without it.

\subsection{Hypothesis 1 - Reduced Workload}

\begin{figure}[!b]
    \centering \textbf{(a)}
    \includegraphics[width=.99 \columnwidth]{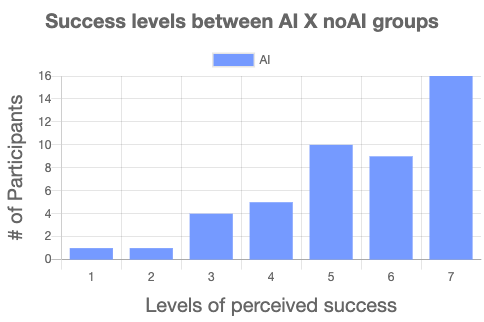}
     \centering\textbf{ (b)}
    \includegraphics[width=.99 \columnwidth]{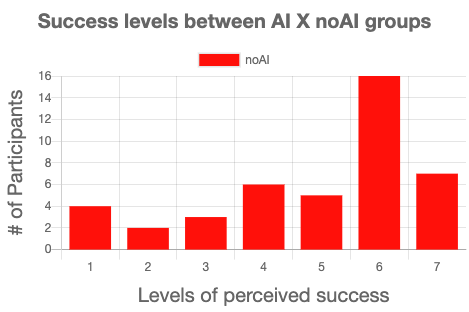}
    \caption{Success levels reported by the respondents in the AI group \textbf{(a)} and non-AI group \textbf{(b)}.}
    \label{fig:success_graph}
\end{figure}

We hypothesized that the recommender system would reduce the amount of workload required to accomplish the goal of designing \textit{Space Invaders}. Our tests showed statistical significance in four out of five questions of the NASA-TLX questionnaire by applying an one sided Wilcoxon-Whitney test. Users in the AI group reported low levels of mental effort (p $\approx$ 0.0003) and insecurity (p $\approx$ 0.0002). They also reported they didn't feel the task was rushed (p $\approx$ 0.0008) and that it required someone to work hard to accomplish what was required in the task (p $\approx 0.002$). The only dimension of the NASA-TLX questionnaire for what we could not find statistical significance was perceived success (p $\approx$ 0.07). By taking a look at the histograms, we can see that the AI group were more confident to report success. Also, they almost did not report a complete fail (see Figure \ref{fig:success_graph}(a)). In opposite, participants in the noAI group were not that confident on report their own success. The report of complete fail was also easily noticeable in this group (see Figure \ref{fig:success_graph}(b)). In general, the analysis of the workload dimension of this study showed that the presence of the recommender system is able to reduce the amount of perceived effort by the users.  

\subsection{Hypothesis 2 - Increased Affect}
We hypothesized that the recommender system would increase the level of computational affect. We applied the PANAS questionnaire and we obtained a statistical significance for the positive and negative scores by using an one-sided Wilcoxon-Whitney test. The experience was more positive for the AI group (p $\approx$ 2.2e-16), and it was more negative for the noAI group (p $\approx$ 1.437e-10). After applying Bonferroni correction we could not see statistical significance for individual dimensions of the questionnaire. However it is worth to note that by analyzing the means for the positive and negative dimensions of the questionnaire, users in the AI group showed better levels of affect in twelve (12) of them. In particular, ``interest', ``excited', ``inspired', ``attentive', ``active', ``distressed', ``upset', ``guilty', ``hostile', ``irritable', ``ashamed', and ``nervous' where the ones when the AI group took advantage. ``Strong', ``enthusiastic', ``proud', ``alert', and ``jittery' were the ones the noAI group got better scores. For ``scared' and ``determined', the means were too close to draw any conclusion. When we analyzed their scores in the questions in which they were asked about their ``upsetness', ``irritability', and ``distressed' levels, we could see that these dimensions are the ones with the biggest variance in this study and they show the AI group faced less discomfort during the experiment. Both groups showed similar levels of 'proudness' with a slight advantage for the noAI group. It raised an hypothesis that the presence of an AI system performing half (or more) of the task can remove the sense of proud in an individual. In general, the AI group reported lower levels of negative effect, and for positive affect the results of both groups were closer, with a small advantage for those in the AI group (see Figure \ref{panas}).

\begin{figure}[!b]
\centering    \includegraphics[width=.99 \columnwidth]{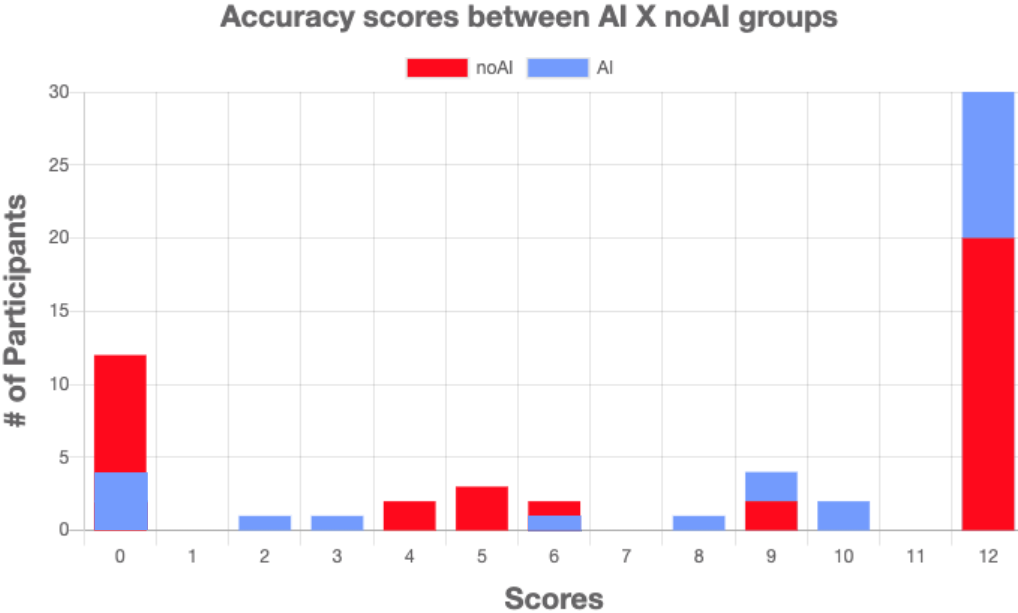}
\caption{Accuracy scores for the two groups. We can see that users in the AI group performed better than the ones in the noAI group. 30 of the users in the AI group delivered their tasks with maximum accuracy (12 points) against 20 from the noAI group. A complete fail (0 points) was less common in the AI group as well. 4 against 12 total fails from the noAI goup.}
\label{fig:game_catalog}
\end{figure}

\subsection{Hypothesis 3 - Increased accuracy}
For evaluating the accuracy, we asked the participants to submit their games after done with the task (either by their own or by time-limit). We designed a program to read their game descriptions and attribute points to it. The program contained all the rules required to run the game exactly as required in the document they received to guide them over the experiment. The rules were divided in two sets, \textit{Sprite} rules and \textit{Interaction} rules to match the VGDL game descriptions sets evaluated in this study. One point was attributed for each rule generated corrected by the users. No half points were attributed, or the rule was totally correct or it would not be enough to get a score. Before running the score evaluation, we ran all the games submitted. Those that could not run got a score of zero (0). From the forty-five (45) submissions of the AI group only four could not run. From the forty-two (42) submissions of the noAI group, twelve of them were not running. After this initial stage, we executed the grader software over the entries of the two groups. Then with all the scores available we ran an one-sided t-test to evaluate the hypothesis that the recommender system would increase the accuracy of the participants. The result showed statistical significance (p $\approx$ 0.004574). Thirty (30) out of Forty-five (45) submissions from the AI-group got the maximum score (12 points) against twenty (20) out of Forty-two (42) submissions in the noAI group. 67\% (AI) against 47\% (noAI). In the AI group, excluding the submissions with the maximum scores and the zeroing ones, only two entries got very low scores (2 and 3), most of the others got scores of 8, 9, and 10 points. The noAI group did not have so low scores after excluding zeroing and maximum scores entries, however they could not get close to the ideal (12) and reached average values like 4, 5, and 6 points. In general, by looking at the final results we saw that most of the errors for the two groups (however with more occurrences in the noAI group) were happening in the interaction set. Or they were missing interactions or using the incorrect sprites to missing them, applying interactions that were not required also was a common mistake presented. For our surprise, we saw two entries in the AI group that even provided the termination set of the game, i.e the conditions that define if a gameplay session results in a win or a lose state. Just for the record, no bonus points were given and, of course, the termination set was not used for evaluation in any case.

\subsection{Hypothesis 4 - Increased Self-Efficacy}
We hypothesized that the presence of the recommender system would increase the participant's Self-Efficacy. However, after analyzing that by applying a one-side Wilcoxon-Whitney test, we could not find statistical significance. For all the questions of the computer self-efficacy scale, just one of them reached a value that could be significant (p $\approx$ 0.047), however not considered after applying Bonferoni correction. The question asked if the participant would be able to perform the task if they never had used a similar product before. Participants' from the AI group were more eager to agree than participants' from the noAI group. This was the only question whose difference of means between the two groups was greater than 1.6. All the others questions had mean difference between 0.27 and 0.82. As the computer self-efficacy scale does not have a method of a evaluate the whole experience as PANAS for example, and because all the questions got results too close for both groups, we do not have how to conclude which of the two groups performed better for this particular (self-efficacy) evaluation. 

\subsection{Qualitative Analysis}
We decided to include an open field where the users would be free to report anything they wanted about the whole experience. It was not mandatory and appeared after the forms previously discussed. 40 people submitted their comments, 22 from the noAI group and 18 from the AI one.

We categorized them in to effort and positive/negative impact by analyzing their inputs using the online free trial of the Atlas.ti software. We could use this qualitative approach to support our previous findings. Users in the AI group were more positive about the experience, basically their speech are all categorized as of positive impact and low effort. One of the participants stated how easy the whole experience was by saying that \textit{``the game dev kit itself makes it super easy to build a game because it tells you what you can choose and based on what you have chosen, it will tell you the possible interactions that can happen between them."}. Another user stated that the system is self explanatory, `\textit{`I really liked how the interactions were suggested so it was self explanatory and leads one to create the game"}. Finally, users also expressed their contentment with the study by affirming that \textit{``it was exciting and fun to play"} and that \textit{``overall it was a fun and pleasant experience"}.

In the noAI group, users' answers were more often categorized as indicating negative impact and high effort. Some of them were complaining about the task's time, \textit{``it requires a little bit more time to watch videos and design level as well. I could complete everything except for the level design."}. Others complained about bugs they found in the tool, \textit{``The application broke part of the way through so I couldn't finish the task. I got as far as making the enemies, but the bombs wouldn't go down properly and I couldn't even see the enemy sprite. Then I made some more enemies exactly as the video showed it(down to the sprite), and those wouldn't show up."}. Even when they expressed positive reactions, they were followed by problems they faced during their experience: \textit{``I really liked the UI of the tool but I had a lot of trouble with the interactions."}.

In general, we saw that users in the AI group would report better experiences and even enjoyment to some extent because the automatic procedures saved them time and effort to learn and even master UI commands. By contrary, participants in the noAI group had to worry about all the procedures to perform the task since they do not had any kind of automatic assistance. 

\subsection{Conclusion}
In this paper we evaluated Pitako, a recommender system for assisting novice game designers, built on the Cicero AI-driven game design assistant. It provides recommendations based on frequent itemset data mining algorithms. Designers get the suggestions while design their games. Their choices tune the system and it is up to them to explore common choices and design clones with small changes, or getting recommendations that lead them to try something new.
Because this tool offers components already created and tends to avoid users effort in design everything from scratch, we hypothesized that such a tool would decrease workload (H1), Increase computational affect (H2), Increase accuracy (H3), and finally, increase self-efficacy (H4). 

We recruited 87 participants and divided them in two groups. We asked them to design the game \textit{Space Invaders}. One group executed the task with Pitako and the other group without it. 
Our results found with statistical significance that the presence of the recommender system decreases the perceived users' workload, increases their computational affect, and increases their accuracy. No statistical significance was found about the users' self - efficacy. 

Computer affect is in particular an interesting way to push this work forward. We found statiscal significancy that the participants' in the AI group had a more positive experience as a whole. However, for particular sub-dimensions of computational affect we could not see (with statistical significance) how participants' got influenced by the AI presence. One of them showed that participants' in the noAI group felt more proud in accomplishing the task. Does that mean that the procedural automatic content suggested (or found) by an AI reduces the proudness level of the user? This is still an open question. Participants also gave us their impressions, that we could categorize by analyzing their free-text answers. The AI group reported a more pleasant experience while the noAI group reported their frustration. The presence of a recommender system in the AI group allowed the participants to keep their focus (almost) entirely on the task, while the noAI group had to learn and remind UI commands that exposed them to more mistakes and difficulties in accomplishing the task.  
We encourage more studies and evaluation of AI-game design assistants with these dimensions in mind: workload, affect, accuracy, and self-efficacy. We are particularly interested in seeing how the experience will change the participants perception when they need to be exposed to the tool for long periods of time, needs to design games of different complexities, and test the tool in both ways (with and without Pitako). 


\section*{Acknowledgements}
 To Conselho Nacional de Desenvolvimento Cient\'ifico e Tecnol\'ogico (CNPQ), Science without Borders scholarship 202859/2015-0. And to our interns: Katherine LosCalzo, Katalina Park, and ZhongHeng Li.

\bibliographystyle{aaai}
\bibliography{sample}
\end{document}